\documentclass[10pt,twocolumn,letterpaper]{article}

\usepackage{iccv}

\usepackage{algorithm}
\usepackage{algpseudocode}

\definecolor{iccvblue}{rgb}{0.21,0.49,0.74}
\usepackage[pagebackref,breaklinks,colorlinks,allcolors=iccvblue]{hyperref}

\def\paperID{8} 
\def\confName{ICCV}
\def\confYear{2025}

\title{DISC-GAN: Disentangling Style and Content for Cluster-Specific Synthetic Underwater Image Generation}

\author{
Sneha Varur \and 
Anirudh R Hanchinamani \and 
Tarun S Bagewadi \and 
Uma Mudenagudi \and 
Chaitra D Desai \and 
Sujata C \and 
Padmashree Desai \and 
Sumit Meharwade \and 
KLE Technological University, Hubballi, Karnataka, India\\
{\tt\small \{
sneha.varur, 
01fe22bcs264, 
01fe22bcs151, 
uma\}@kletech.ac.in}\\
{\tt\small \{
chaitra.desai, 
sujata\_c, 
padmashri,
01fe21bci032\}@kletech.ac.in}
}
\begin{document}
\maketitle

\begin{figure*}[t]
    \centering
    \includegraphics[width=\textwidth]{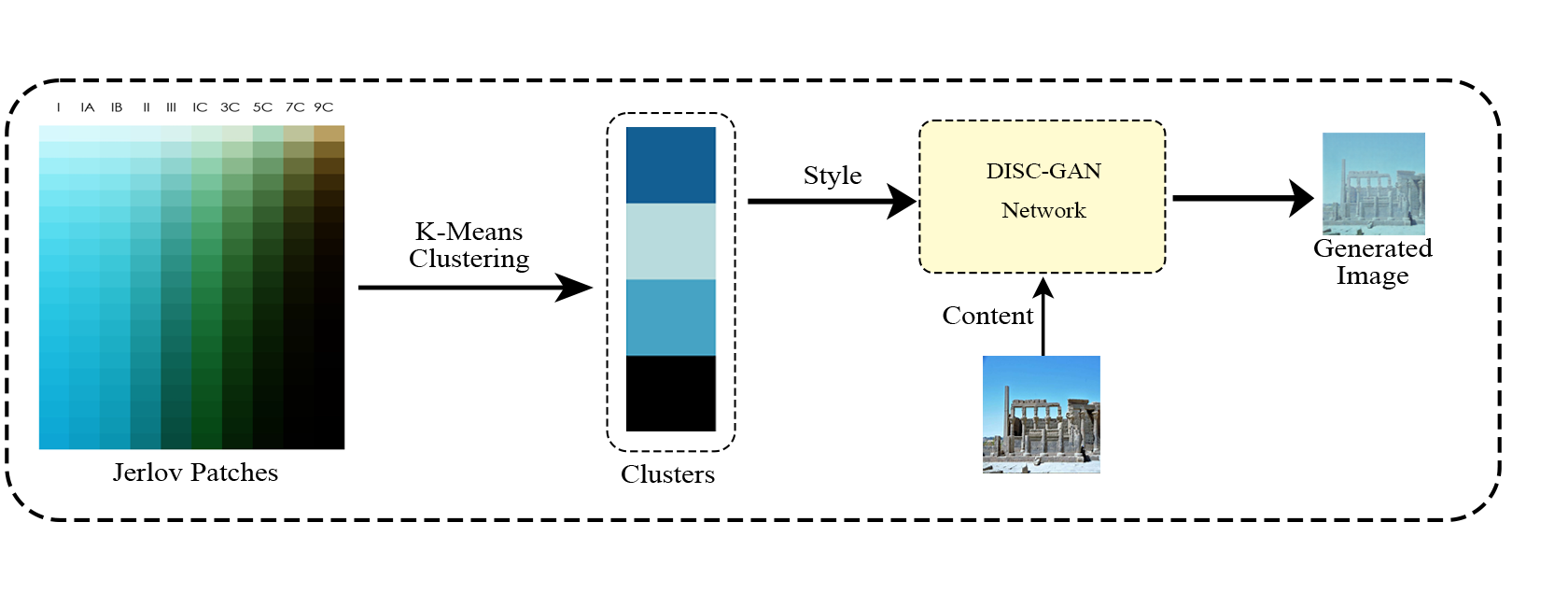}
    \caption{\textbf{The high-level design of our proposed DISC-GAN framework}. DISC-GAN is trained using clusters obtained from image patches derived from the Jerlov water classification scheme, originally introduced by Jerlov in "Marine Optics" (1976 \cite{jerlov1976marine}). The visualization of the Jerlov water types and the associated example water patch are reproduced from Desai et al., "Realistic Synthetic Underwater Image Generation with Image Formation Model" (2024 \cite{desai2022rsuigmdataset}). This visualization originally appeared in Akkaynak and Treibitz, "What Is the Space of Attenuation Coefficients in Underwater Computer Vision?" (2017 \cite{Akkaynak_2017_CVPR}).}
    \label{fig:high-level-design}
\end{figure*}

\begin{abstract}
In this paper, we propose a novel framework, Disentangled Style-Content GAN (DISC-GAN), which integrates style-content disentanglement with a cluster-specific training strategy towards photorealistic underwater image synthesis. The quality of synthetic underwater images is challenged by optical distortions due to phenomena such as color attenuation and turbidity. These phenomena are represented by distinct stylistic variations across different waterbodies, such as changes in tint and haze. While generative models are well-suited to capture complex patterns, they often lack the ability to model the non-uniform stylistic conditions of diverse underwater environments. To address these challenges, we employ K-means clustering to partition a dataset into style-specific domains. We use separate encoders to get latent spaces for style and content; we further integrate these latent representations via Adaptive Instance Normalization (AdaIN) and decode the result to produce the final synthetic image. The model is trained independently on each style cluster to preserve domain-specific characteristics. Our framework demonstrates state-of-the-art performance, obtaining a Structural Similarity Index (SSIM) of 0.9012, an average Peak Signal-to-Noise Ratio (PSNR) of 32.5118 dB, and a Fréchet Inception Distance (FID) of 13.3728.
\end{abstract}    
\section{Introduction}
\label{sec:introduction}

In this paper, we propose a novel generative framework for underwater image synthesis, termed as Disentangled Style-Content GAN (DISC-GAN). The aim is to generate photorealistic underwater images by modeling the diverse optical conditions of different waterbodies. Synthesizing underwater images is challenging because of the complex optical interactions that take place during image formation. Unlike atmospheric environments, underwater scenes are affected by severe, depth-dependent light attenuation and absorption, which cause a rapid loss of longer wavelengths and result in a dominant blue-green color distortion \cite{jerlov1976marine, chiang2012underwater, cong2024comprehensive}. Simultaneously, suspended particles scatter light in forward and backward directions, creating image blur and a veiling haze that reduces contrast and obscures distant objects \cite{ancuti2012enhancing, akkaynak2019sea}. These degradations affect the visual quality of images and limit the performance of tasks such as marine monitoring, exploration, and robotics.

Researchers in this domain address underwater synthesis and restoration in two ways: physics-based and learning-based. Physics-based models, for instance, have proposed using monocular depth estimation as a prior to guide restoration \cite{desai2023depthcue, zhao2021unpaired}, fusion-based techniques using color and contrast priors \cite{ancuti2012enhancing}, or models based on wavelength compensation and dehazing \cite{chiang2012underwater, cong2024comprehensive, akkaynak2019sea}. However, these methods often rely on hand-crafted features, which limits their generalizability across diverse underwater scenes.

Alternatively, learning-based methods train neural networks on real-world or synthetic datasets to directly map between domains. Methods such as WaterGAN \cite{li2017watergan} demonstrated how to generate synthetic data by simulating light attenuation on in-air RGB-D pairs, while other generative approaches have used variational models \cite{dai2024single} or cycle-consistency loss for unpaired enhancement \cite{fabbri2018enhancing, zhu2017cyclegan}. An extensive body of work has explored GANs for this task, focusing on multi-domain translation \cite{bakht2024mula}, texture consistency \cite{wei2019underwater}, perceptual losses \cite{peng2020end}, and lightweight networks for real-time performance \cite{islam2020fast, shen2022lightweight}. However, many of these methods do not explicitly model the fine-grained stylistic differences between waterbody types, making them less controllable and limiting their realism for domain-specific applications.

Hybrid methods and advanced GAN architectures have attempted to merge physical priors with learning-based techniques to address these limitations. Style transfer using Adaptive Instance Normalization (AdaIN) \cite{huang2017adain} and perceptual losses \cite{johnson2016perceptual} have enabled greater structural preservation. Disentangled GANs, such as the architecture proposed by Kazemi et al. \cite{kazemi2018scgan} and others \cite{zhang2017separating}, have shown how to learn latent representations of content and style independently. Recent works have also incorporated depth-guidance \cite{cao2021underwater, desai2023depthcue}, camera-awareness \cite{li2021uiec2net}, or dual-domain learning \cite{li2017watergan, zhang2019dual} to improve results. However, the explicit partitioning of the underwater domain into distinct stylistic clusters, inspired by physical water characteristics, remains underexplored within these learning frameworks.

Towards this, we propose DISC-GAN, a framework that includes a style clustering module to partition the underwater domain before training a style-content disentangled GAN. Our work adopts the Realistic Synthetic Underwater Image Generation with Image Formation Model (RSUIGM) dataset \cite{desai2022rsuigmdataset}, which provides physically accurate stylistic variations across different Jerlov water types \cite{jerlov1976marine}. The usage of this dataset ensures that our learned style clusters are grounded in real-world optical properties. The proposed two-phase pipeline is shown in Figure \ref{fig:high-level-design}. We combine physics-informed data partitioning with a learning-based synthesis model and make the following contributions:

\begin{itemize}
    \item We propose a method to partition underwater images into distinct style domains via K-means clustering on a feature space combining color histograms and mean depth values, inspired by Jerlov water types.
    \item We introduce a novel, cluster-specific synthesis framework that generates realistic underwater images by training a separate GAN on each style domain to prevent style leakage.
    \item We develop an effective style-content disentangled GAN, which uses AdaIN to preserve content structure from terrestrial images while transferring style from the target underwater domain, validated by high SSIM, PSNR and low FID scores.
\end{itemize}

In Section~\hyperref[sec:methodology]{II}, we describe our synthesis framework, outlining each component from style clustering to image synthesis. Section~\hyperref[sec:implementation]{III} provides the implementation details of our model. In Section~\hyperref[sec:results]{IV}, we discuss the results of our experiments. Finally, Section~\hyperref[sec:conclusion]{V} concludes the paper and suggests future work.

\section{Style Clustering and Image Synthesis}

\label{sec:methodology}

In this section, we propose a novel framework, Disentangled Style-Content GAN (DISC-GAN), for cluster-specific underwater image generation that integrates domain partitioning with a deep learning model. The synthesis framework has three modules corresponding to three phases:
\begin{itemize}
    \item dataset preprocessing and style clustering,
    \item style-content feature disentanglement and fusion, and
    \item learning-based synthesis,
\end{itemize}
as shown in Figure~\ref{fig:final_arch}. In Section~\hyperref[subsec:twoone]{2.1}, we explain the datasets and our preprocessing pipeline. In Section~\hyperref[subsec:twotwo]{2.2}, we detail the methodology for partitioning the dataset into distinct style clusters. In Section~\hyperref[subsec:twothree]{2.3}, we focus on the disentangled synthesis architecture and the composite loss function. Finally, in Section~\hyperref[subsec:twofour]{2.4}, we provide an overview of the training process.

\subsection{Dataset and Preprocessing}
\label{subsec:twoone}

\begin{figure}
    \centering    \includegraphics[width=1\linewidth]{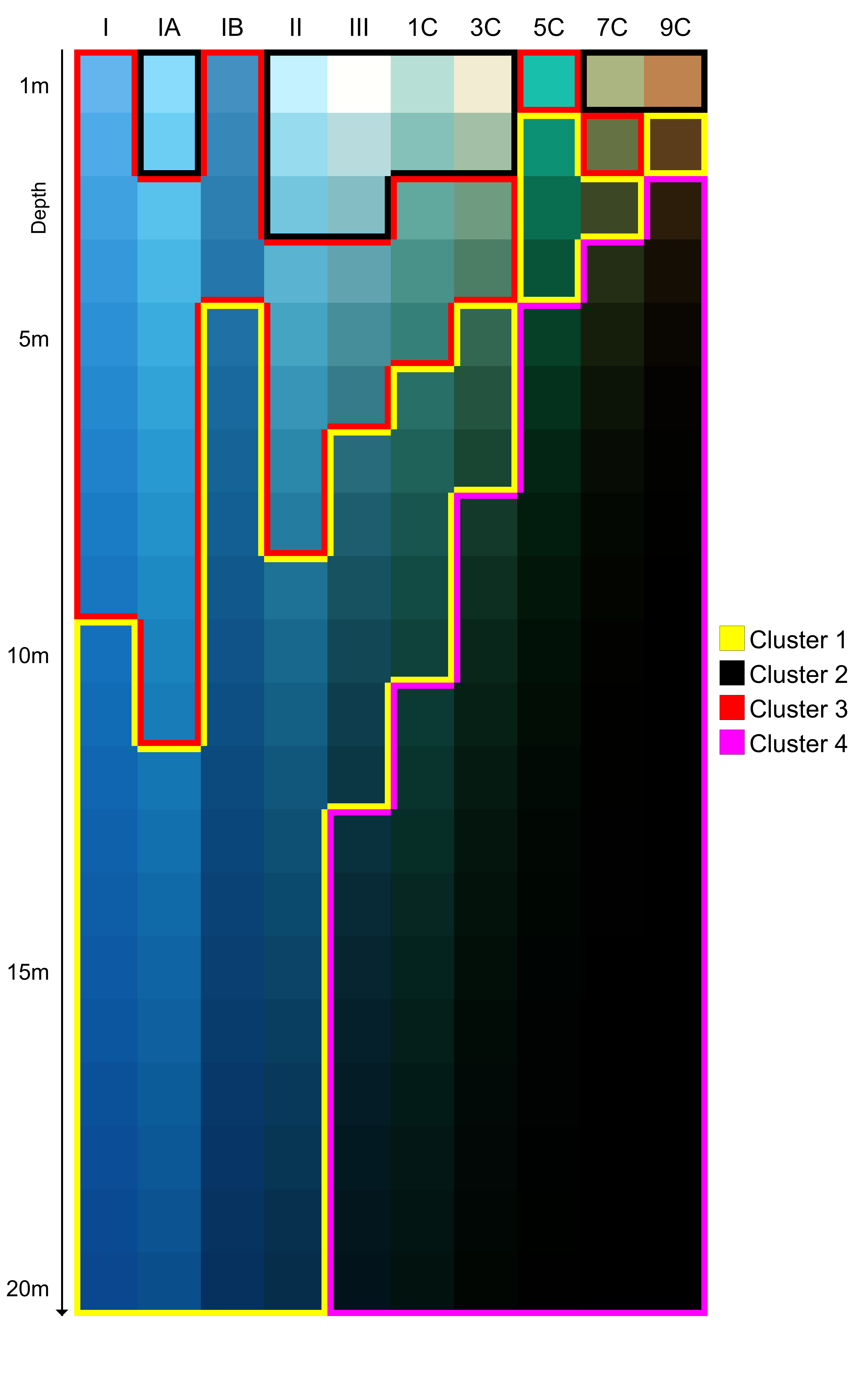}
    \caption{Example patches from the RSUIGM dataset used for style clustering.}
    \label{fig:style_patches}
\end{figure}

The Parameter Estimator module is designed to predict the parameters that govern underwater image formation, enabling physics-informed underwater image restoration. For style modeling and training, we utilize the RSUIGM (Realistic Synthetic Underwater Image Generation Model) dataset \cite{desai2022rsuigmdataset}, a synthetic benchmark generated by simulating underwater image formation under various water conditions. Its consistency with a physically grounded image formation model makes it ideal for training our style-content disentanglement framework. Each synthetically degraded image is paired with a clean reference and a corresponding depth map, allowing for supervised learning. In our pipeline, clean images serve as content inputs, while the distorted underwater versions are used to model style vectors.

For training our proposed DISC-GAN framework, we specifically employ the RSUIGM dataset due to its physically inspired modeling of underwater light propagation. RSUIGM is generated using an image formation model that accounts for both downwelling depth ($z$) and line-of-sight distance ($d$), thereby incorporating realistic spatial variations in light attenuation and scattering. The dataset consists of 6000 synthetic underwater images rendered across diverse ocean bed and coastal bed scenes, spanning the full range of Jerlov water types. These images simulate characteristic degradations such as haze, blur, color attenuation, and tint, making RSUIGM particularly suitable for training data-driven generative models. In our work, RSUIGM serves as the ground-truth reference for guiding DISC-GAN synthesis and ensuring that generated images retain the statistical and perceptual characteristics of real underwater imagery.

In addition to RSUIGM, we use the SUID (Standard Underwater Image Dataset) as a source of clean terrestrial images for content extraction. The dataset provides high-resolution images with diverse structural features, which helps the content encoder generalize across complex scene geometries. To ensure consistency, all input images are resized to a 256x256 resolution and augmented via horizontal flipping. Corresponding depth maps are downsampled to match, and RGB histograms are extracted from the RSUIGM style images for the clustering phase. For training and validation, the dataset is split into a 80:20 ratio.

\subsection{Style Domain Clustering}
\label{subsec:twotwo}
To model the diverse appearances of underwater scenes, we first partition the RSUIGM dataset into visually coherent style domains. This process is designed to group images based on both color and physical depth properties, inspired by the classification of oceanic water into Jerlov water types \cite{jerlov1976marine}. The physical basis for this clustering comes from the RSUIGM image formation model \cite{desai2022rsuigmdataset}, which simulates degradation using the Beer-Lambert law:
\begin{equation}
\label{eq:image_formation}
I_c(x) = J_c(x) \cdot e^{-K_c \cdot d(x)} + B_c \cdot \left(1 - e^{-K_c \cdot d(x)}\right)
\end{equation}

Here, \(I_c(x)\) is the observed intensity, \(J_c(x)\) is the clean radiance, \(K_c\) is the attenuation coefficient, \(d(x)\) is depth, and \(B_c\) is background light. To capture these degradation patterns, each image is converted into a feature vector composed of its RGB histogram and normalized mean depth.

\begin{figure}[htbp]
    \centering
    \includegraphics[width=1\linewidth]{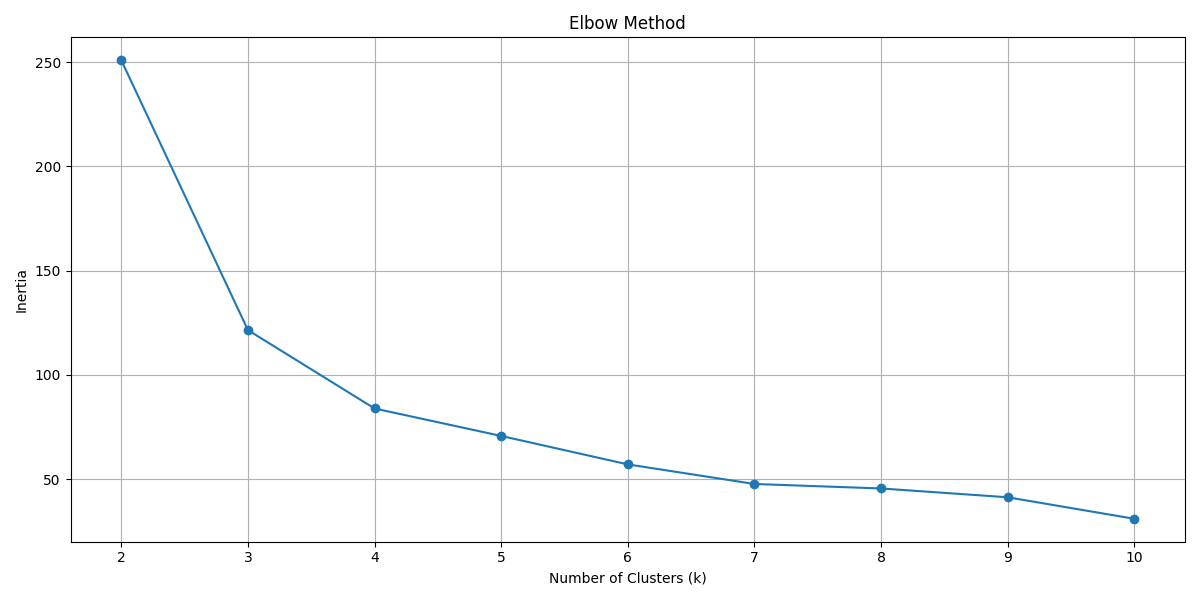}
    \caption{The Elbow Method plot for K-means clustering on the 200 classes of Jerlov considering the rbg values for each class. The "elbow" point at k=4 suggests it is the optimal number of clusters.}
    \label{fig:elbow_method}
\end{figure}

\begin{figure*}[htbp]
    \centering
    \begin{subfigure}[b]{0.24\textwidth}
        \centering
        \includegraphics[width=\linewidth, height=0.35\textheight, keepaspectratio]{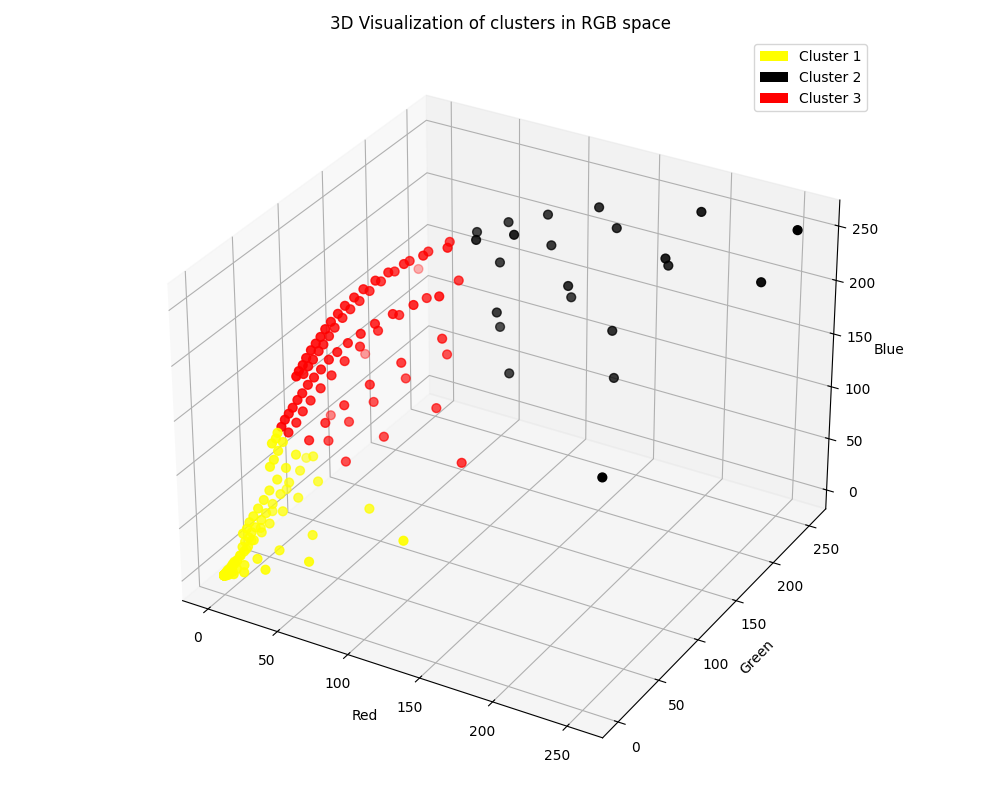}
        \caption{3D, k=3}
        \label{fig:3d_k3}
    \end{subfigure}
    \hfill
    \begin{subfigure}[b]{0.24\textwidth}
        \centering
        \includegraphics[width=\linewidth, height=0.25\textheight, keepaspectratio]{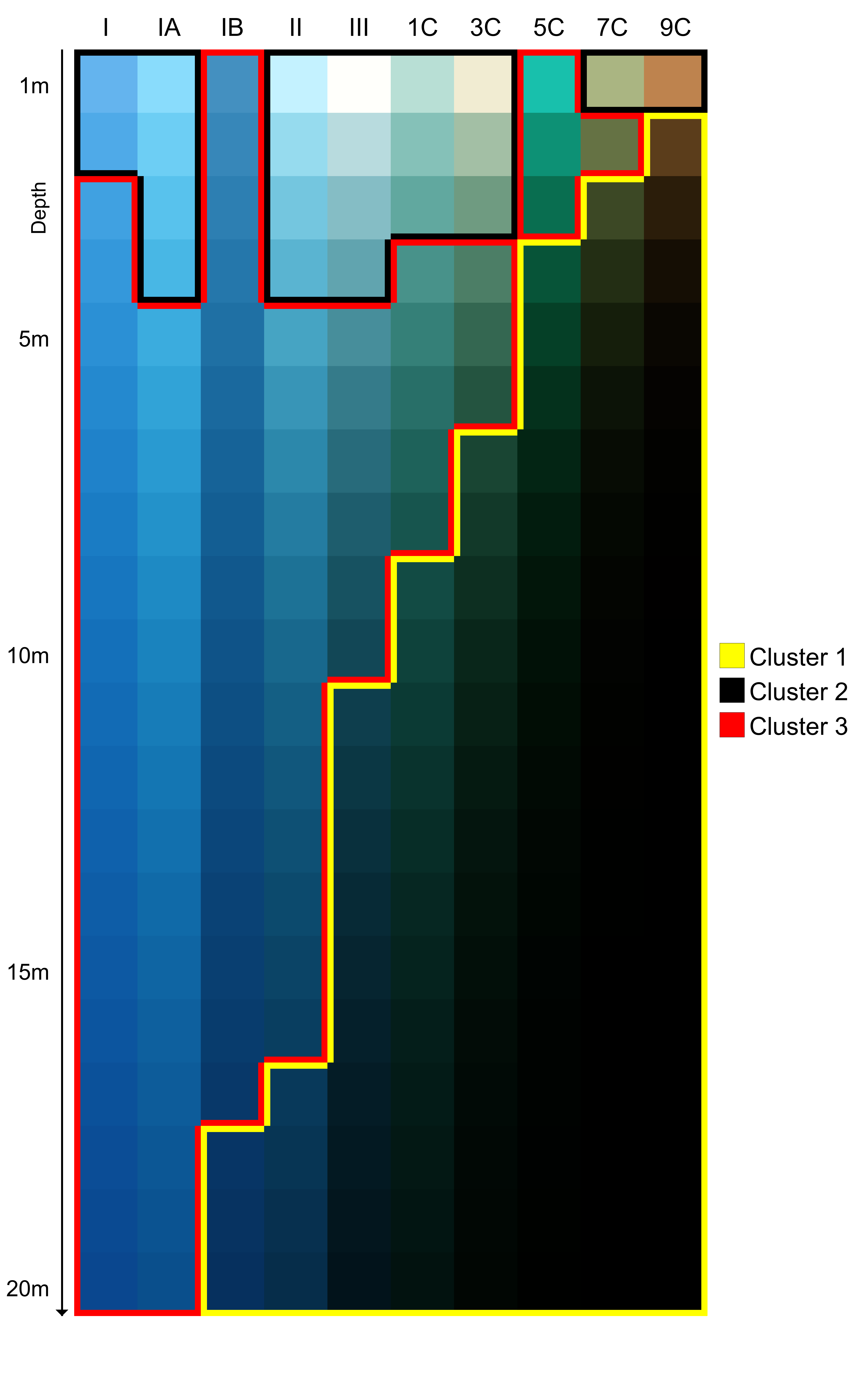}
        \caption{Patches, k=3}
        \label{fig:patches_k3}
    \end{subfigure}
    \hfill
    \begin{subfigure}[b]{0.24\textwidth}
        \centering
        \includegraphics[width=\linewidth, height=0.25\textheight, keepaspectratio]{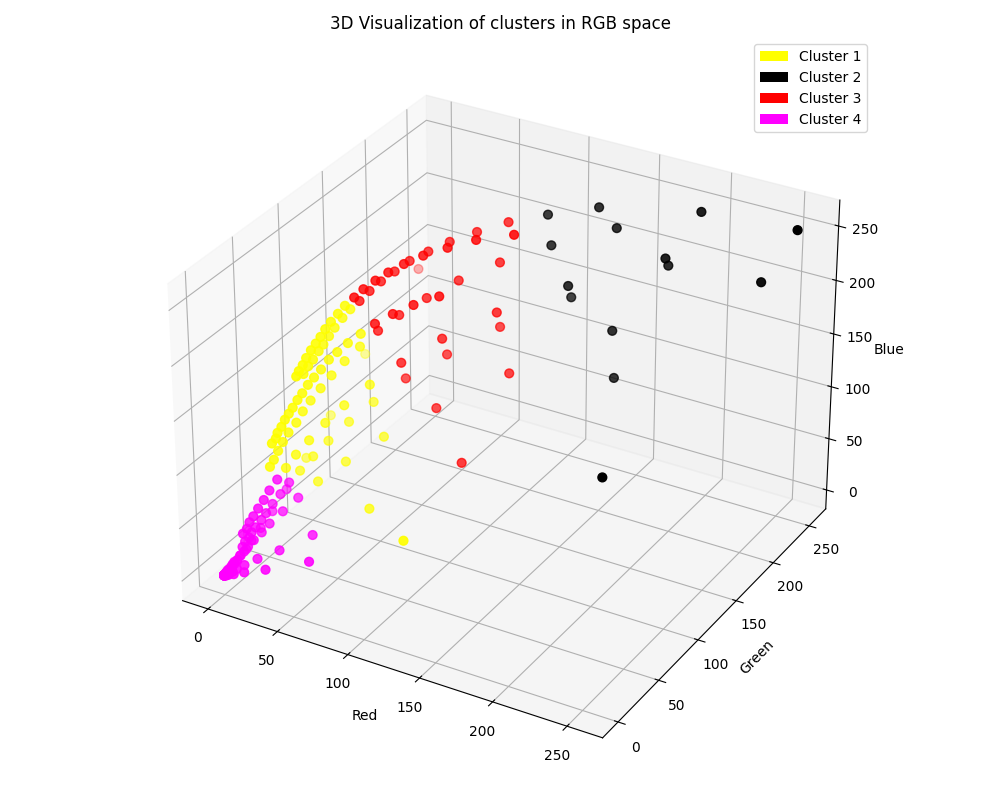}
        \caption{3D, k=4 (Optimal)}
        \label{fig:3d_k4}
    \end{subfigure}
    \hfill
    \begin{subfigure}[b]{0.24\textwidth}
        \centering
        \includegraphics[width=\linewidth, height=0.25\textheight, keepaspectratio]{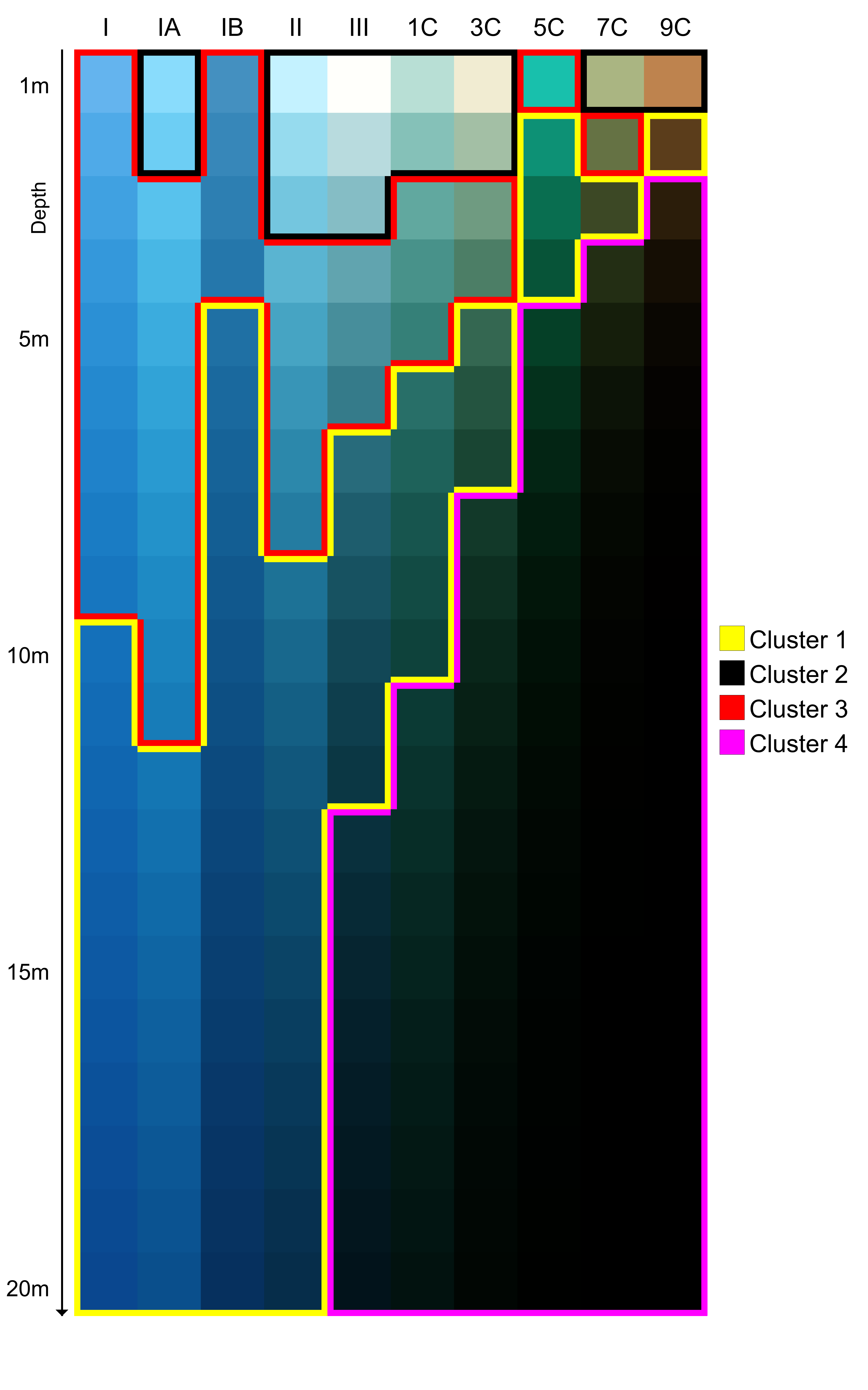}
        \caption{Patches, k=4 (Optimal)}
        \label{fig:patches_k4}
    \end{subfigure}

    \vspace{0.2cm} 

    \begin{subfigure}[b]{0.24\textwidth}
        \centering
        \includegraphics[width=\linewidth, height=0.25\textheight, keepaspectratio]{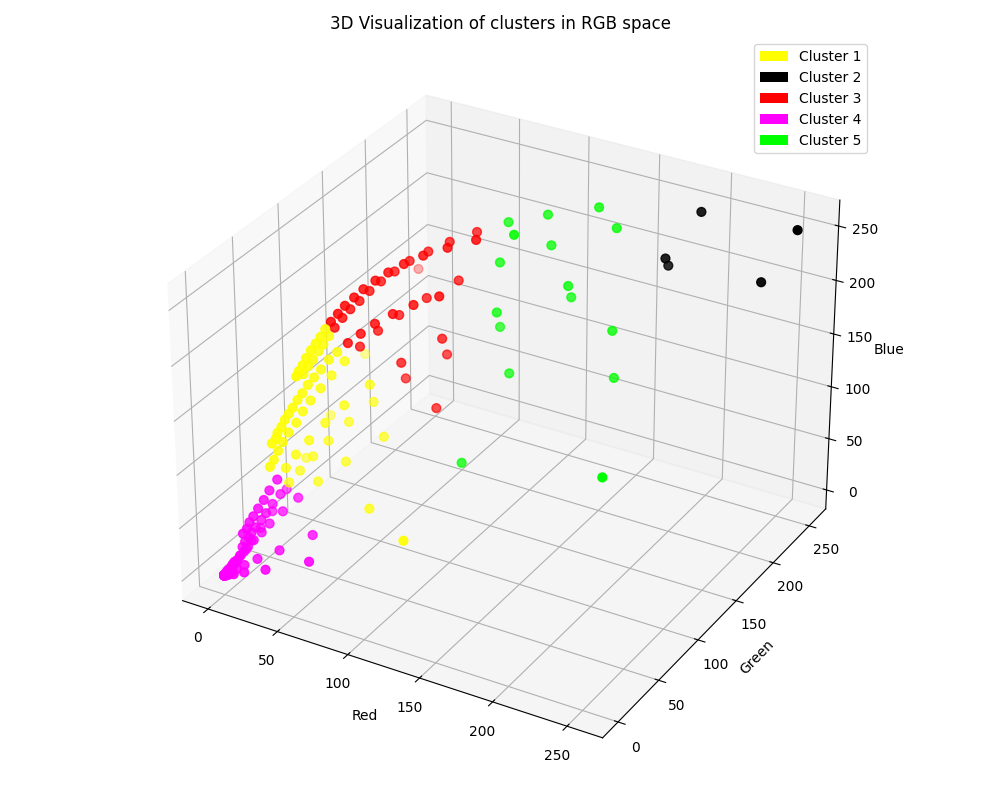}
        \caption{3D, k=5}
        \label{fig:3d_k5}
    \end{subfigure}
    \hfill
    \begin{subfigure}[b]{0.24\textwidth}
        \centering
        \includegraphics[width=\linewidth, height=0.25\textheight, keepaspectratio]{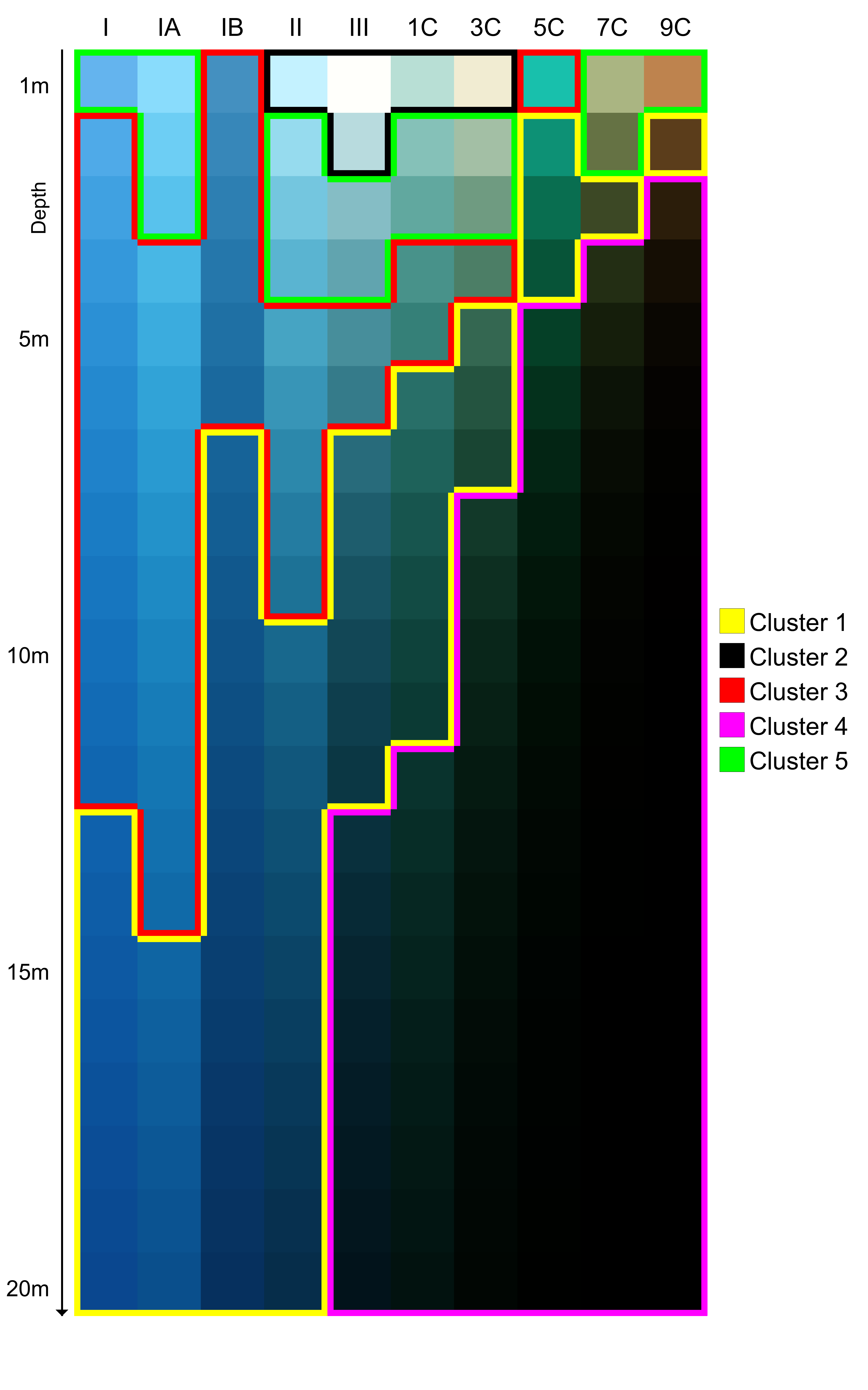}
        \caption{Patches, k=5}
        \label{fig:patches_k5}
    \end{subfigure}
    \hfill
    \begin{subfigure}[b]{0.24\textwidth}
        \centering
        \includegraphics[width=\linewidth, height=0.25\textheight, keepaspectratio]{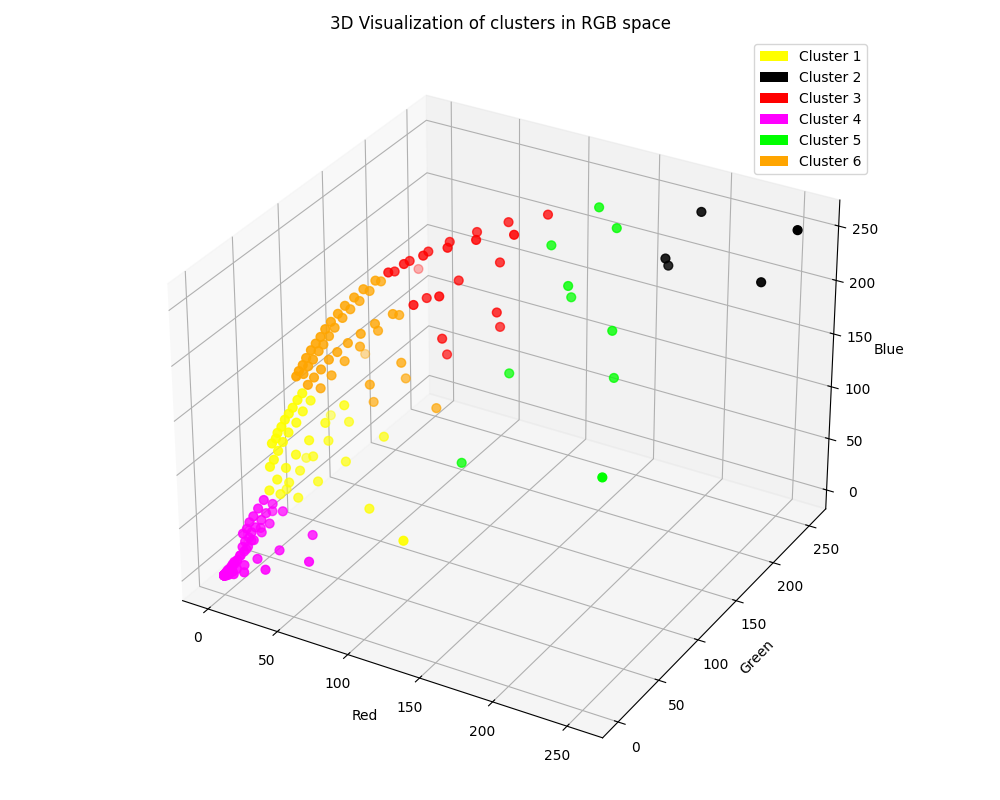}
        \caption{3D, k=6}
        \label{fig:3d_k6}
    \end{subfigure}
    \hfill
    \begin{subfigure}[b]{0.24\textwidth}
        \centering
        \includegraphics[width=\linewidth, height=0.25\textheight, keepaspectratio]{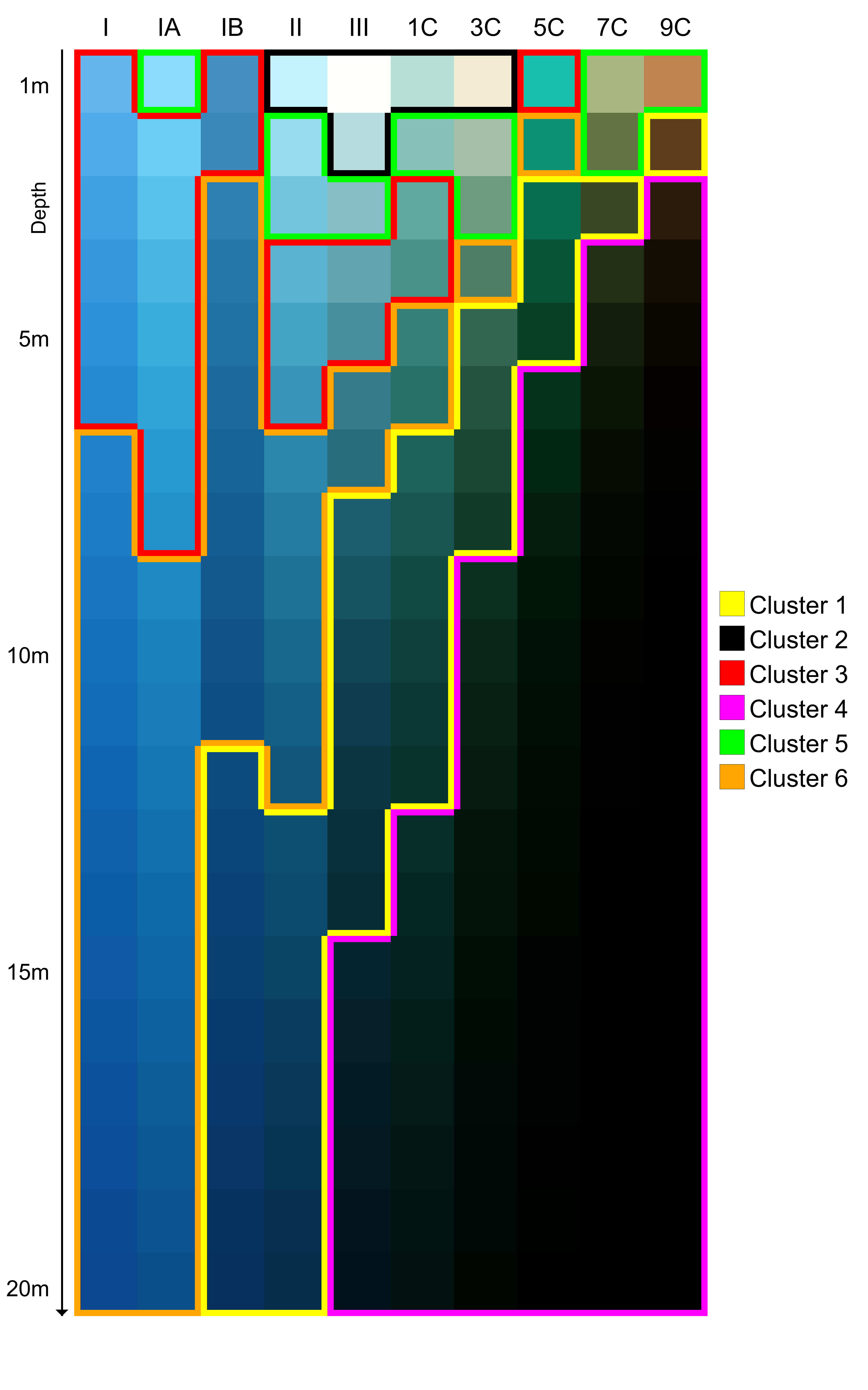}
        \caption{Patches, k=6}
        \label{fig:patches_k6}
    \end{subfigure}

    \caption{Visual comparison of K-means clustering results for k=3, 4, 5, and 6. The top row compares k=3 and k=4 (optimal), while the bottom row compares k=5 and k=6. Each pair shows the 3D RGB plot and its corresponding clustered jerlov classes. The results for k=4 show the most visually coherent and distinct style separation.}
    \label{fig:cluster_comparison}
\end{figure*}

We then apply K-means clustering to partition the data by minimizing the intra-cluster variance according to the objective:
\begin{equation}
\label{eq:kmeans}
\arg\min_{C} \sum_{i=1}^{k} \sum_{x \in C_i} \| x - \mu_i \|^2
\end{equation}
where \(C_i\) is the \(i^{th}\) cluster and \(\mu_i\) is its centroid. The optimal number of clusters was determined to be \(k=4\) using the Elbow Method. This yields four dominant underwater styles: \textbf{blue}, \textbf{light-blue}, \textbf{dark-blue}, and \textbf{black}, as shown in Figure \ref{fig:elbow_method}. The clear separation of these clusters in 3D RGB space, shown in Figure \ref{fig:cluster_comparison}, validates that each group represents a distinct and visually coherent style domain.

To visually validate this choice, we compared the clustering results for \(k \in \{3, 4, 5, 6\}\), as shown in Figure \ref{fig:cluster_comparison}. This comparison includes both the 3D RGB space visualization and the corresponding clustered image patches for each value of \(k\). For \(k=3\), distinct style domains appear to be merged into single, less coherent groups. Conversely, for \(k=5\) and \(k=6\), visually similar styles are unnecessarily split, leading to over-segmentation and redundancy. The configuration for \(k=4\) (Figures \ref{fig:cluster_comparison}(c) and \ref{fig:cluster_comparison}(d)) provides the most meaningful separation, yielding four dominant and visually distinct underwater styles, which we label as \textbf{blue}, \textbf{light-blue}, \textbf{dark-blue}, and \textbf{black}. The clear separation of these clusters in 3D RGB space validates that each group represents a distinct and coherent style domain.

\subsection{Disentangled Synthesis Framework}
\label{subsec:twothree}
\begin{figure*}
    \centering
    \includegraphics[width=1\linewidth]{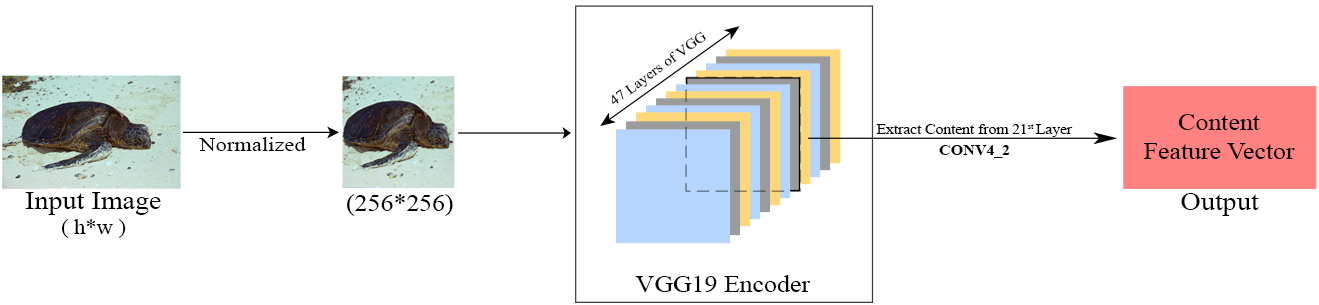}
    \caption{A content feature vector is extracted from the input image using the conv4\_2 layer of a VGG19 encoder, preserving high-level structural information.}
    \label{fig:content_extraction}
\end{figure*}

The final stage of the framework is the synthesis pipeline, which transforms latent style and content features into a photorealistic image, drawing inspiration from the original SC-GAN architecture \cite{kazemi2018scgan}. Given a clean content image \(I_c\) and a style reference \(I_s\) from a target cluster, the framework first uses separate encoders. The content encoder \(\mathcal{E}_{content}\) produces a latent tensor \(z_c\):
\begin{equation}
z_c = \mathcal{E}_{content}(I_c)
\end{equation}
Simultaneously, the style encoder \(\mathcal{E}_{style}\) extracts a style vector \(z_s\):
\begin{equation}
z_s = \mathcal{E}_{style}(I_s)
\end{equation}
These features are fused within the generator \(G\) using Adaptive Instance Normalization (AdaIN) \cite{huang2017adain}, which aligns the feature statistics to inject the style without altering content structure. The generator then synthesizes the final stylized output \(\hat{I}\):
\begin{equation}
\hat{I} = G(z_c, z_s)
\end{equation}
The operation yields a final stylized image \(\hat{I}\) reconstructed from disentangled and physics-guided latent features. To ensure the generator produces high-fidelity outputs, we use a composite loss function that combines L1 reconstruction loss with an adversarial loss. The final generator loss \(\mathcal{L}_G\) is given by:
\begin{equation}
\mathcal{L}_G = \mathcal{L}_{L1(\hat{y},y)} + \lambda \cdot \mathcal{L}_{2}
\label{eq:generator_loss}
\end{equation}
Here, the adversarial term \(\mathcal{L}_{G_{\text{GAN}}}\) enforces perceptual realism, while the L1 term \(\mathcal{L}_{G_{\text{L1}}}\) encourages structural preservation. The hyperparameter \(\lambda\) balances these two objectives. This composite loss approach follows standard practice in conditional image synthesis, where an adversarial loss improves visual realism \cite{goodfellow2014generative} and a pixel-wise loss like L1 maintains structural fidelity to the target \cite{isola2017pix2pix}.

\subsection{Training Strategy}
\label{subsec:twofour}
The training process of the model happens in two phases, as outlined in Algorithm \ref{alg:discgan}.
\begin{enumerate}
    \item \textbf{Style Domain Partitioning:} The RSUIGM dataset is first partitioned into four distinct style clusters using the K-means algorithm based on color and depth features. This ensures that the model can learn waterbody-specific styles in a controlled manner.
    \item \textbf{Cluster-Specific GAN Training:} The generator and discriminator are then trained in an alternating fashion. For each training step, a content image from SUID and a style reference from one of the four RSUIGM clusters are used. The composite loss is back-propagated to update the generator and discriminator. The entire GAN is trained independently for each of the four style clusters to prevent style leakage and enhance control. This sequential procedure ensures that the synthesis model benefits from a stable and meaningful feature space shaped by the physics-informed clustering.
\end{enumerate}

\begin{algorithm}[htbp]
\caption{Cluster-Specific Synthetic Underwater Image Generation using DISC-GAN}
\label{alg:discgan}
\begin{algorithmic}[1]
\State \textbf{Initialize:} Generator $\theta_G$, Discriminator $\theta_D$, dataset $\mathcal{D}$, clusters $\mathcal{C}$, encoders $\mathcal{E}_{content}, \mathcal{E}_{style}$
\State \textbf{Phase I: Clustering}
\State Partition dataset $\mathcal{D}$ into $k=4$ style clusters $\mathcal{C}$ using K-means.
\Statex
\State \textbf{Phase II: Training}
\For{each style cluster $c \in \mathcal{C}$}
    \For{each content image $I_{cont}$ and style image $I_{style} \in c$}
        \State $z_c \gets \mathcal{E}_{content}(I_{cont})$
        \State $z_s \gets \mathcal{E}_{style}(I_{style})$
        \State $\hat{I} \gets G(z_c, z_s)$
        \State Calculate $\mathcal{L}_G$ using Eq. \ref{eq:generator_loss}
        \State Update $\theta_G$ and $\theta_D$ via gradient descent
    \EndFor
\EndFor
\Statex
\State \textbf{Inference:}
\State Given content image $I_c$ and target cluster $c$
\State Select random style image $I_s$ from cluster $c$
\State $\hat{I} \gets G(\mathcal{E}_{content}(I_c), \mathcal{E}_{style}(I_s))$
\State \Return $\hat{I}$
\end{algorithmic}
\end{algorithm}
\section{Implementation Details}
\label{sec:implementation}

\begin{figure*}[htbp]
    \centering
    \includegraphics[width=1\linewidth]{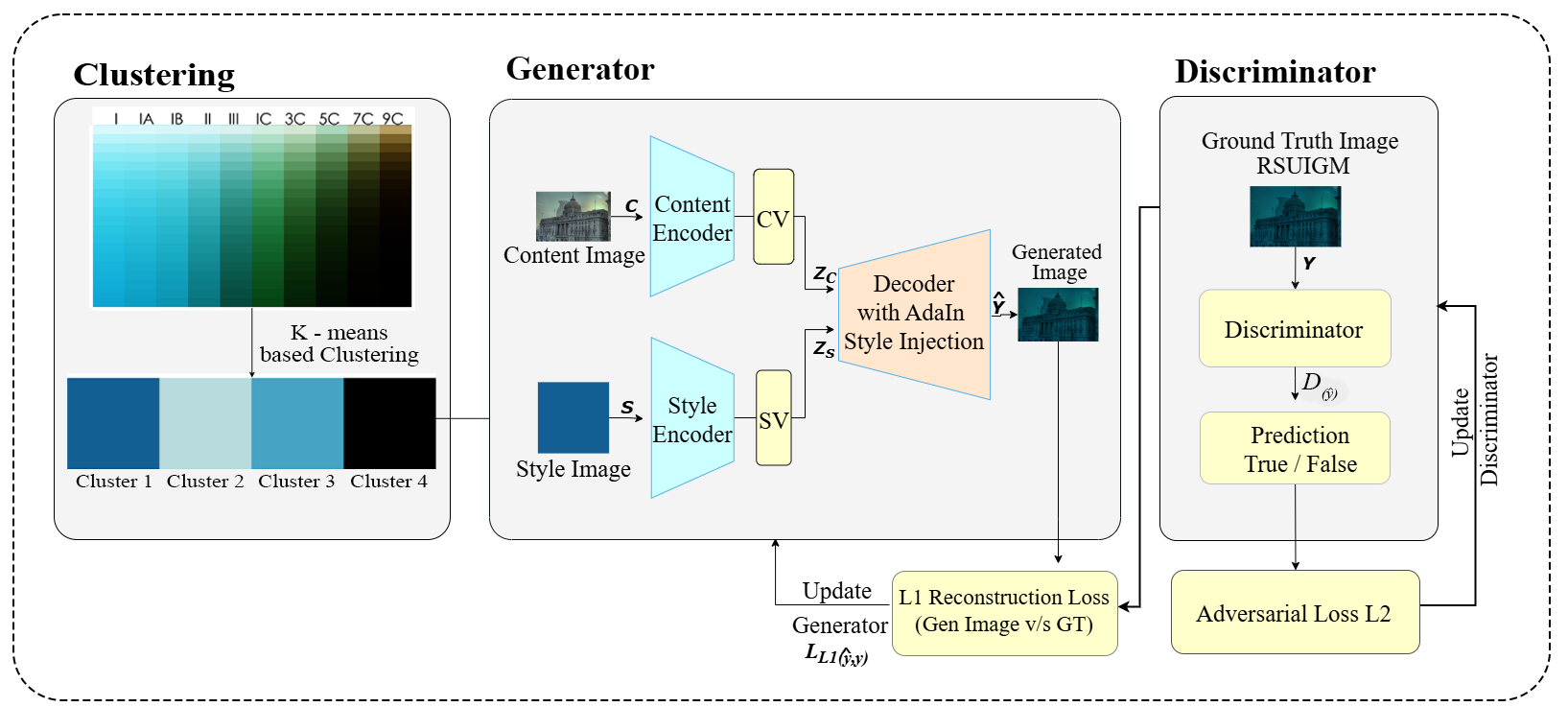}
    \caption{The complete architecture of DISC-GAN. The pipeline shows the parallel encoding of content and style images, feature fusion via Adaptive Instance Normalization (AdaIN), the generative decoder, and the PatchGAN discriminator that guides training using a composite loss.}
    \label{fig:final_arch}
\end{figure*}

The end-to-end architecture of the proposed Disentangled Style-Content GAN (DISC-GAN) is illustrated in Figure \ref{fig:final_arch}. The framework is composed of three primary modules: a content encoder, a style encoder, and a generator with a corresponding discriminator. These components are designed to learn disentangled mappings for controlled and realistic underwater image synthesis.

\subsection{Network Architecture}
Our framework utilizes encoders built upon the VGG19 network \cite{simonyan2014very}, pretrained on ImageNet, to extract high-quality content and style representations.

\textbf{Content and Style Encoders.} To achieve disentanglement, feature extraction is stratified. The content encoder uses the deeper \texttt{relu4\_2} layer of the VGG19 network to capture high-level semantic and structural information from the clean input image \cite{johnson2016perceptual, gatys2015neural}. In contrast, the style encoder is designed to extract global appearance statistics. It computes Gram matrices from the feature maps of shallower layers (\texttt{relu1\_1}, \texttt{relu2\_1}, \texttt{relu3\_1}) to effectively model the low-level texture and color tint from the underwater style reference image \cite{gatys2015neural}.

\textbf{Generator and Discriminator.} The generator receives the extracted content features and fuses them with the style representation using Adaptive Instance Normalization (AdaIN) layers \cite{huang2017adain}. This fused tensor is then processed through a series of residual blocks and transposed convolutions to decode the final, stylized output image. For adversarial training, we employ a PatchGAN discriminator \cite{isola2017pix2pix}, which evaluates realism by classifying 70x70 overlapping patches of the image. This approach provides localized feedback, better preserving high-frequency details.

\subsection{Training Parameters}
The DISC-GAN framework is trained end-to-end by optimizing the composite loss function described in Equation \ref{eq:generator_loss}, which combines an L1 reconstruction loss with an L2 adversarial loss. The final objective for the generator is to minimize:
\begin{equation}
\mathcal{L}_G = \lambda_{\text{rec}} \|I_s - \hat{I}\|_1 + \lambda_{\text{adv}} \|D(\hat{I}) - 1\|_2^2
\end{equation}
where \(I_s\) is the target style image, \(\hat{I}\) is the generated image, and \(D(\cdot)\) is the discriminator's output. The hyperparameters \(\lambda_{\text{rec}}\) and \(\lambda_{\text{adv}}\) balance structural fidelity with perceptual realism.

The training is optimized using the Adam optimizer with a learning rate of \(2 \times 10^{-4}\) and momentum parameters \(\beta_1 = 0.5\) and \(\beta_2 = 0.999\). The models were implemented using PyTorch and trained for 100 epochs for each of the four style clusters on an NVIDIA Tesla V100 GPU. The full training and inference procedure is formalized in Algorithm \ref{alg:discgan}.

\section{Results}
\label{sec:results}

In this section, we present the evaluation of our proposed DISC-GAN framework on the RSUIGM dataset \cite{desai2022rsuigmdataset}. We report results on the synthesis quality across the four pre-defined style clusters and analyze the visual fidelity of the generated images using full-reference image quality metrics. We also compare the performance of our data-driven method against the physics-informed principles used to create the ground-truth dataset through both quantitative and qualitative analysis.

\subsection{Quantitative Analysis}
We use a learning-based framework, DISC-GAN, for the synthesis of underwater images. To evaluate its performance, we use three standard quantitative metrics: the Structural Similarity Index Measure (SSIM), the Peak Signal-to-Noise Ratio (PSNR), and the Fréchet Inception Distance (FID) \cite{yu2021fid}. SSIM and PSNR assess structural and pixel-level similarity to the ground truth, while FID measures the distributional similarity between real and generated images. The Fréchet Inception Distance is given by:
\begin{equation}
\text{FID} = \|\mu_r - \mu_g\|^2 + \text{Tr}\left( \Sigma_r + \Sigma_g - 2(\Sigma_r \Sigma_g)^{1/2} \right)
\end{equation}
where $(\mu_r, \Sigma_r)$ and $(\mu_g, \Sigma_g)$ are the mean and covariance of features from real and generated images, respectively. A lower FID score indicates better perceptual quality.

The model was evaluated for its ability to generate high-fidelity images for each of the four style clusters. The quantitative scores are reported in Table \ref{tab:metrics}. The model demonstrates robust performance, achieving high SSIM and PSNR values alongside low FID scores. Notably, Cluster 1 (blue) achieved an SSIM of 0.9012 and a PSNR of 32.5118, indicating strong structural preservation. The low FID scores across all clusters, such as 3.8576 for Cluster 4 (black), confirm that the generated images closely match the statistical distribution of the real underwater scenes in the RSUIGM dataset.

\begin{table}[h]
\centering
\caption{Quantitative evaluation of DISC-GAN across the four style clusters using SSIM, PSNR, and FID metrics.}
\label{tab:metrics}
\begin{tabular}{|c|c|c|c|c|}
\hline
\textbf{Metric} & \textbf{Blue} & \textbf{Light-Blue} & \textbf{Dark-Blue} & \textbf{Black} \\
\hline
SSIM & 0.9012 & 0.8107 & 0.7551 & 0.7212\\
PSNR & 32.5118 & 27.8125 & 31.7876 & 40.8871\\
FID & 8.3728 & 7.6894 & 7.0021 & 3.8576\\
\hline
\end{tabular}
\end{table}

\subsection{Qualitative Analysis}
\begin{figure*}[htbp]
    \centering
    \includegraphics[width=0.8\linewidth]{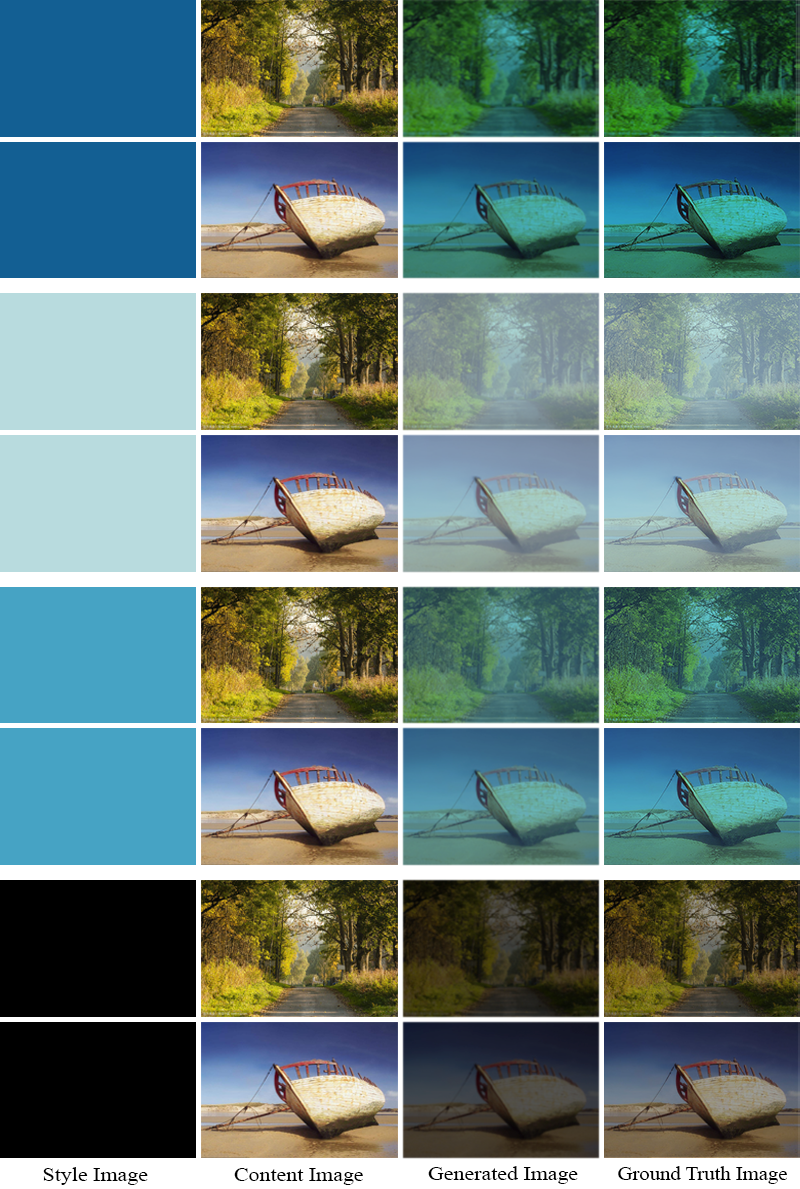}
    \caption{Qualitative results of the DISC-GAN framework. Each row shows a different content image, while each column represents a different target style cluster (Blue, Light-Blue, Dark-Blue, and Black). The content remains consistent while the underwater style changes across columns.}
    \label{fig:qual_results}
\end{figure*}

We provide a qualitative comparison of the synthesized images for each style cluster in Figure \ref{fig:qual_results}. We observe that DISC-GAN successfully applies distinct underwater styles to a variety of clean content images. The model capably adapts tint, haze, and illumination to match the target cluster while preserving the structural integrity of the input content. The generated outputs are visually consistent with real underwater scenes and show minimal structural artifacts. This modular generation capability validates the effectiveness of our cluster-wise training approach and the successful disentanglement of style and content.

\subsection{Analysis of Data-Driven Synthesis}
A significant outcome of our work is the demonstration that a purely data-driven approach can achieve a level of realism comparable to that of physics-informed methodologies. The RSUIGM dataset, used here as ground truth, was generated with a model cued by the physical principles of underwater light attenuation. Our DISC-GAN framework, without being explicitly programmed with these physical constraints, has learned to synthesize images that are quantitatively and qualitatively similar. This result validates that deep learning models, when structured for effective disentanglement, can internalize complex physical phenomena from data alone, presenting a powerful and flexible alternative to traditional physics-based rendering. Although spatial variation is not explicitly modeled in our pipeline, the GAN leverages the RSUIGM dataset to implicitly capture and reproduce these variations, thereby contributing to the realism of the generated underwater images.

\section{Conclusion}
\label{sec:conclusion}

In this paper, we have proposed a novel framework for the synthesis of underwater images that integrates style-domain partitioning and deep learning. The system is structured into three modules: K-means clustering to define style domains, disentangled feature learning using VGG19 encoders, and image generation via a decoder with Adaptive Instance Normalization. Unlike traditional methods that treat the underwater environment as visually homogenous, our approach benefits from cluster-specific training, which provides fine-grained control over the generated appearance. We have demonstrated the qualitative and quantitative effectiveness of the model by synthesizing realistic underwater images and evaluating performance using metrics such as PSNR, SSIM, and FID.

To improve its performance and expand its applicability, future work may explore the integration of temporal consistency for video synthesis, which would enable real-time simulation for underwater video feeds. The framework could be further enhanced by incorporating attention mechanisms or depth-aware style modulation to achieve more precise blending. Expanding the clustering mechanism to consider additional water quality metrics such as turbidity or salinity could also increase realism for specialized marine domains. These strategies can enhance the model's adaptability and accuracy, strengthening its applications in fields such as autonomous navigation, data augmentation for object detection, and environmental modeling.
{
    \small
    \bibliographystyle{ieeenat_fullname}
    \bibliography{main}

\begin{thebibliography}{30}
\providecommand{\natexlab}[1]{#1}
\providecommand{\url}[1]{\texttt{#1}}
\expandafter\ifx\csname urlstyle\endcsname\relax
  \providecommand{\doi}[1]{doi: #1}\else
  \providecommand{\doi}{doi: \begingroup \urlstyle{rm}\Url}\fi

\bibitem[Akkaynak and Treibitz(2019)]{akkaynak2019sea}
Derya Akkaynak and Tali Treibitz.
\newblock Sea-thru: A method for removing water from underwater images.
\newblock In \emph{Proceedings of the IEEE/CVF conference on computer vision
  and pattern recognition}, pages 1682--1691, 2019.

\bibitem[Akkaynak et~al.(2017)Akkaynak, Treibitz, Shlesinger, Loya, Tamir, and
  Iluz]{Akkaynak_2017_CVPR}
Derya Akkaynak, Tali Treibitz, Tom Shlesinger, Yossi Loya, Raz Tamir, and David
  Iluz.
\newblock What is the space of attenuation coefficients in underwater computer
  vision?
\newblock In \emph{Proceedings of the IEEE Conference on Computer Vision and
  Pattern Recognition (CVPR)}, 2017.

\bibitem[Ancuti et~al.(2012)Ancuti, Ancuti, Haber, and
  Bekaert]{ancuti2012enhancing}
Cosmin Ancuti, Codruta~Orniana Ancuti, Tom Haber, and Philippe Bekaert.
\newblock Enhancing underwater images and videos by fusion.
\newblock In \emph{2012 IEEE conference on computer vision and pattern
  recognition}, pages 81--88. IEEE, 2012.

\bibitem[Bakht et~al.(2024)Bakht, Jia, Din, Akram, Saoud, Seneviratne, Lin, He,
  and Hussain]{bakht2024mula}
Ahsan~B Bakht, Zikai Jia, Muhayy~Ud Din, Waseem Akram, Lyes~Saad Saoud, Lakmal
  Seneviratne, Defu Lin, Shaoming He, and Irfan Hussain.
\newblock Mula-gan: Multi-level attention gan for enhanced underwater
  visibility.
\newblock \emph{Ecological Informatics}, 81:\penalty0 102631, 2024.

\bibitem[Chaurasia and Chhikara(2024)]{cao2021underwater}
Dhiraj Chaurasia and Prateek Chhikara.
\newblock Sea-pix-gan: Underwater image enhancement using adversarial neural
  network.
\newblock \emph{Journal of Visual Communication and Image Representation},
  98:\penalty0 104021, 2024.

\bibitem[Chiang and Chen(2011)]{chiang2012underwater}
John~Y Chiang and Ying-Ching Chen.
\newblock Underwater image enhancement by wavelength compensation and dehazing.
\newblock \emph{IEEE transactions on image processing}, 21\penalty0
  (4):\penalty0 1756--1769, 2011.

\bibitem[Cong et~al.(2024)Cong, Zhao, Gui, Hou, and Tao]{cong2024comprehensive}
Xiaofeng Cong, Yu Zhao, Jie Gui, Junming Hou, and Dacheng Tao.
\newblock A comprehensive survey on underwater image enhancement based on deep
  learning.
\newblock \emph{arXiv preprint arXiv:2405.19684}, 2024.

\bibitem[Dai and Lin(2024)]{dai2024single}
Chenggang Dai and Mingxing Lin.
\newblock Single underwater image restoration using variational framework
  guided by imaging model with noise.
\newblock \emph{IEEE Access}, 12:\penalty0 82427--82442, 2024.

\bibitem[Desai et~al.(2023)Desai, Benur, Tabib, Patil, and
  Mudenagudi]{desai2023depthcue}
Chaitra Desai, Sujay Benur, Ramesh~Ashok Tabib, Ujwala Patil, and Uma
  Mudenagudi.
\newblock Depthcue: Restoration of underwater images using monocular depth as a
  clue.
\newblock In \emph{Proceedings of the IEEE/CVF Winter Conference on
  Applications of Computer Vision (WACV) Workshops}, pages 196--205, 2023.

\bibitem[Desai et~al.(2024)Desai, Benur, Patil, and
  Mudenagudi]{desai2022rsuigmdataset}
Chaitra Desai, Sujay Benur, Ujwala Patil, and Uma Mudenagudi.
\newblock Rsuigm: Realistic synthetic underwater image generation with image
  formation model.
\newblock \emph{ACM Transactions on Multimedia Computing, Communications and
  Applications}, 21\penalty0 (1):\penalty0 1--22, 2024.

\bibitem[Fabbri et~al.(2018)Fabbri, Islam, and Sattar]{fabbri2018enhancing}
Cameron Fabbri, Md~Jahidul Islam, and Junaed Sattar.
\newblock Enhancing underwater imagery using generative adversarial networks.
\newblock In \emph{2018 IEEE international conference on robotics and
  automation (ICRA)}, pages 7159--7165. IEEE, 2018.

\bibitem[Gatys et~al.(2015)Gatys, Ecker, and Bethge]{gatys2015neural}
Leon~A Gatys, Alexander~S Ecker, and Matthias Bethge.
\newblock A neural algorithm of artistic style.
\newblock \emph{arXiv preprint arXiv:1508.06576}, 2015.

\bibitem[Goodfellow et~al.(2020)Goodfellow, Pouget-Abadie, Mirza, Xu,
  Warde-Farley, Ozair, Courville, and Bengio]{goodfellow2014generative}
Ian Goodfellow, Jean Pouget-Abadie, Mehdi Mirza, Bing Xu, David Warde-Farley,
  Sherjil Ozair, Aaron Courville, and Yoshua Bengio.
\newblock Generative adversarial networks.
\newblock \emph{Communications of the ACM}, 63\penalty0 (11):\penalty0
  139--144, 2020.

\bibitem[Huang and Belongie(2017)]{huang2017adain}
Xun Huang and Serge Belongie.
\newblock Arbitrary style transfer in real-time with adaptive instance
  normalization.
\newblock In \emph{Proceedings of the IEEE International Conference on Computer
  Vision (ICCV)}, 2017.

\bibitem[Islam et~al.(2020)Islam, Xia, and Sattar]{islam2020fast}
Md~Jahidul Islam, Youya Xia, and Junaed Sattar.
\newblock Fast underwater image enhancement for improved visual perception.
\newblock \emph{IEEE robotics and automation letters}, 5\penalty0 (2):\penalty0
  3227--3234, 2020.

\bibitem[Isola et~al.(2017)Isola, Zhu, Zhou, and Efros]{isola2017pix2pix}
Phillip Isola, Jun-Yan Zhu, Tinghui Zhou, and Alexei~A Efros.
\newblock Image-to-image translation with conditional adversarial networks.
\newblock In \emph{Proceedings of the IEEE conference on computer vision and
  pattern recognition}, pages 1125--1134, 2017.

\bibitem[Jerlov(1976)]{jerlov1976marine}
Nils~Gunnar Jerlov.
\newblock \emph{Marine optics}.
\newblock Elsevier, 1976.

\bibitem[Johnson et~al.(2016)Johnson, Alahi, and
  Fei-Fei]{johnson2016perceptual}
Justin Johnson, Alexandre Alahi, and Li Fei-Fei.
\newblock Perceptual losses for real-time style transfer and super-resolution.
\newblock In \emph{European conference on computer vision}, pages 694--711.
  Springer, 2016.

\bibitem[Kazemi et~al.(2019)Kazemi, Iranmanesh, and Nasrabadi]{kazemi2018scgan}
Hadi Kazemi, Seyed~Mehdi Iranmanesh, and Nasser Nasrabadi.
\newblock Style and content disentanglement in generative adversarial networks.
\newblock In \emph{2019 IEEE Winter Conference on Applications of Computer
  Vision (WACV)}, pages 848--856. IEEE, 2019.

\bibitem[Li et~al.(2017)Li, Skinner, Eustice, and
  Johnson-Roberson]{li2017watergan}
Jie Li, Katherine~A Skinner, Ryan~M Eustice, and Matthew Johnson-Roberson.
\newblock Watergan: Unsupervised generative network to enable real-time color
  correction of monocular underwater images.
\newblock \emph{IEEE Robotics and Automation letters}, 3\penalty0 (1):\penalty0
  387--394, 2017.

\bibitem[Simonyan and Zisserman(2014)]{simonyan2014very}
Karen Simonyan and Andrew Zisserman.
\newblock Very deep convolutional networks for large-scale image recognition.
\newblock \emph{arXiv preprint arXiv:1409.1556}, 2014.

\bibitem[Sivaanpu et~al.(2022)Sivaanpu, Priyadarshani, Kokul, and
  Ramanan]{zhang2019dual}
Anparasy Sivaanpu, Rasika Priyadarshani, Thanikasalam Kokul, and Amirthalingam
  Ramanan.
\newblock Underwater image enhancement using dual convolutional neural network
  with skip connections.
\newblock In \emph{2022 2nd International Conference on Advanced Research in
  Computing (ICARC)}, pages 224--229. IEEE, 2022.

\bibitem[Tolie et~al.(2024)Tolie, Ren, Hasan, and Kannan]{li2021uiec2net}
Hamidreza~Farhadi Tolie, Jinchang Ren, Md~Junayed Hasan, and Somasundar Kannan.
\newblock Enhancing underwater situational awareness: Realsense camera
  integration with deep learning for improved depth perception and distance
  measurement.
\newblock In \emph{Artificial Intelligence for Security and Defence
  Applications II}, pages 34--42. SPIE, 2024.

\bibitem[Wu et~al.(2022)Wu, Liu, Lu, Lin, Qin, and Shi]{wei2019underwater}
Junjun Wu, Xilin Liu, Qinghua Lu, Zeqin Lin, Ningwei Qin, and Qingwu Shi.
\newblock Fw-gan: Underwater image enhancement using generative adversarial
  network with multi-scale fusion.
\newblock \emph{Signal Processing: Image Communication}, 109:\penalty0 116855,
  2022.

\bibitem[Yang et~al.(2024)Yang, Xu, Lin, and He]{shen2022lightweight}
Haodong Yang, Jisheng Xu, Zhiliang Lin, and Jianping He.
\newblock Lu2net: a lightweight network for real-time underwater image
  enhancement.
\newblock \emph{arXiv preprint arXiv:2406.14973}, 2024.

\bibitem[Yu and Qin(2023)]{peng2020end}
Yang Yu and Chenfeng Qin.
\newblock An end-to-end underwater-image-enhancement framework based on
  fractional integral retinex and unsupervised autoencoder.
\newblock \emph{Fractal and Fractional}, 7\penalty0 (1):\penalty0 70, 2023.

\bibitem[Yu et~al.(2021)Yu, Zhang, and Deng]{yu2021fid}
Yu Yu, Weibin Zhang, and Yun Deng.
\newblock Frechet inception distance (fid) for evaluating gans.
\newblock \emph{China University of Mining Technology Beijing Graduate School},
  3\penalty0 (11), 2021.

\bibitem[Zhang et~al.(2018)Zhang, Zhang, and Cai]{zhang2017separating}
Yexun Zhang, Ya Zhang, and Wenbin Cai.
\newblock Separating style and content for generalized style transfer.
\newblock In \emph{Proceedings of the IEEE conference on computer vision and
  pattern recognition}, pages 8447--8455, 2018.

\bibitem[Zhao et~al.(2021)Zhao, Xin, Yu, and Zheng]{zhao2021unpaired}
Qi Zhao, Zhichao Xin, Zhibin Yu, and Bing Zheng.
\newblock Unpaired underwater image synthesis with a disentangled
  representation for underwater depth map prediction.
\newblock \emph{Sensors}, 21\penalty0 (9):\penalty0 3268, 2021.

\bibitem[Zhu et~al.(2017)Zhu, Park, Isola, and Efros]{zhu2017cyclegan}
Jun-Yan Zhu, Taesung Park, Phillip Isola, and Alexei~A Efros.
\newblock Unpaired image-to-image translation using cycle-consistent
  adversarial networks.
\newblock In \emph{Proceedings of the IEEE international conference on computer
  vision}, pages 2223--2232, 2017.

\end{thebibliography}
}

\end{document}